\documentclass[runningheads]{llncs}
\usepackage[hidelinks]{hyperref} % hidelinks to remove url bounding box
\usepackage[T1]{fontenc}
\usepackage{graphicx}
\usepackage{caption}
\usepackage{subcaption}
\usepackage{float}
\usepackage{amsmath}
\usepackage{makecell}
\usepackage{cite}
\usepackage{enumitem}
\setlist{nosep}

\usepackage{orcidlink}

\newcommand{\repthanks}[1]{\textsuperscript{\ref{#1}}}
\makeatletter
\patchcmd{\maketitle}
{\def\thanks}
{\let\repthanks\repthanksunskip\def\thanks}
{}{}
\patchcmd{\@maketitle}
{\def\thanks}
{\let\repthanks\@gobble\def\thanks}
{}{}
\newcommand\repthanksunskip[1]{\unskip{}}
\makeatother

\begin{document}

\title{KidRisk: Benchmark Dataset for Children Dangerous Action Recognition}
\titlerunning{Children Dangerous Action Recognition}

\author{
Minh-Kha Nguyen\thanks{These authors contributed equally to this research. \protect\label{X}}\inst{1, 2} 
\and Trung-Hieu Do\repthanks{X}\inst{1, 2} 
\and Kim Anh Phung\inst{3}\orcidlink{0000-0002-4561-7536} 
\and Thao Thi Phuong Dao\inst{1, 2, 4}\orcidlink{0000-0002-0109-1114}
\and Minh-Triet Tran\inst{1, 2}\orcidlink{0000-0003-3046-3041} 
\and 
Trung-Nghia Le \thanks{Corresponding author.} \inst{1,2}\orcidlink{0000-0002-7363-2610}}

\institute{
University of Science, VNU-HCM, Vietnam 
\and Vietnam National University, Ho Chi Minh City, Vietnam 
\and PrimeLabs LLC, United States\\
\and Thong Nhat Hospital, Ho Chi Minh City, Vietnam\\
\email{\{20120502,20120007\}@student.hcmus.edu.vn, kimaphung@gmail.com},
\email{thao.dao2020@ict.jvn.edu.vn},
\email{\{tmtriet,ltnghia\}@fit.hcmus.edu.vn}}

\authorrunning{M.-K. Nguyen et al.}

\maketitle              % typeset the header of the contribution
\begin{abstract}
Children are naturally energetic, and during their spontaneous activities, they often encounter potentially dangerous situations, especially when lacking parental supervision. Identifying actions that pose risks plays a crucial role in ensuring their safety. This paper build a novel challenging dataset, namely KidRisk, including 2,500 short videos of children's actions and 10,000 images for dangerous action of children. We also introduce a benchmark on our newly constructs dataset and find that traditional deep learning models demonstrated limited effectiveness on these tasks. Therefore, we develop vision-language based baselines with exceptional context understanding of visual information. Our proposed methods achieved an accuracy of 83.53\% in classifying children's actions and 96.14\% in recognizing children's dangerous actions, significantly outperforming traditional approaches. These results confirm that vision-language models are not only feasible but also highly effective in detecting hazardous actions, contributing positively to safeguarding children's safety.

% Action recognition in computer vision is vital for ensuring safety, particularly in child monitoring, where detecting dangerous actions is crucial due to children's curiosity and lack of awareness. Traditional monitoring systems require constant supervision, which is impractical, and existing models struggle with child-related data due to being trained primarily on adult datasets. Additionally, the scarcity of data on dangerous actions in children presents a challenge. However, the development of vision-language models like CLIP, ALIGN, and BLIP offers a promising solution by linking visual data with contextual information, improving the detection of dangerous situations. This paper proposes using the BLIP-2 model for transfer learning and creating a new dataset focused on dangerous actions in children to enhance action recognition and contextual understanding. Experiments demonstrate that integrating BLIP-2 with LSTM achieves 96.1% accuracy in detecting dangerous situations and 83.5% accuracy in action recognition, highlighting the model's effectiveness even with limited data, making it practical for real-world child safety monitoring applications.

\keywords{Action recognition, dangerous action recognition, vision-language model}
\end{abstract}

\section{Introduction}
Action recognition is a crucial research area in computer vision that involves identifying and classifying actions from images or videos. In the context of child monitoring, recognizing dangerous actions is of paramount importance to ensure safety. Children often exhibit curiosity and lack of awareness about their surroundings, leading to potentially hazardous situations such as falls, collisions, or contact with sharp objects. With the advancement of technology, camera systems are becoming increasingly common for monitoring purposes. However, these systems typically require constant human supervision, which can be both demanding and impractical. An intelligent monitoring system can timely detect these actions and alert caregivers to potential risks.

Despite significant advances in action recognition, applying this technology in child monitoring presents several challenges. One of the significant challenges is that existing action recognition models are typically trained on adult datasets, resulting in poor performance when applied to child-related data. Additionally, there is a scarcity of data on dangerous actions in children, making it difficult to train deep learning models effectively. Moreover, many current systems overlook the context and environmental factors, leading to inaccurate assessments of danger levels.

The rapid development of vision-language models has introduced powerful tools for understanding and interpreting complex visual contexts, which is particularly promising for enhancing child safety. Models such as CLIP \cite{radford2021learning}, ALIGN \cite{jia2021scaling}, and BLIP \cite{li2022blip} exemplify this advancement. These models excel in linking visual data with textual descriptions and can interpret and generate nuanced contextual information. We assume that applying such advanced vision-language models to child monitoring may enhance broader contextual understanding of dangerous situations, resulting in enhancing child safety.

This paper aims to propose a promising approach capable of accurately recognizing and analyzing dangerous actions. To this end, we construct a new dataset called KidRisk focused on dangerous actions in children, built up on the InfAct dataset \cite{huang2023posture}. The KidRisk dataset consists of 2,500 short videos of children’s actions and 10,000 images for dangerous action of children. The dataset can capture a range of dangerous actions specific to children, enhancing the model's ability to generalize from limited examples. Our dataset also addresses issues such as the limited number of samples for each action and the imbalance between unsafe and safe labels. We also introduce a benchmark with state-of-the-art action recognition methods. Furthermore, we propose using the BLIP-2 \cite{li2023blip} model for transfer learning to maximize the effectiveness of children dangerous action recognition. Additionally, to further improve the model’s versatility, we utilize zero-shot learning techniques for classifying actions it has not seen during training.

Extensive experiments on the newly constructed dataset demonstrate the efficacy of our proposed method. Specifically, the integration of BLIP-2 with LSTM significantly enhances the performance of action recognition and danger situation detection. With this approach, we achieved a remarkable accuracy of 96.1\% in detecting dangerous situations, showcasing a substantial improvement over traditional models. For action recognition, our method attained an accuracy rate of 83.5\%. These results highlight not only the effectiveness of transfer learning in leveraging pre-trained vision-language models but also its ability to adapt effectively to specific datasets. Our approach shows that, even with limited training data, it is possible to achieve high accuracy and reliability. This underscores the practicality of our solution for real-world applications, particularly in monitoring and ensuring child safety, where accurate and timely detection of dangerous actions is critical.

Our contributions are as follow:
\begin{itemize}
    \item We present a new dataset designed for both action recognition and danger situation recognition in children's activities. This dataset, coupled with a benchmark, tackles real-world challenges such as the scarcity of samples per action and the imbalance between safe and unsafe labels, ensuring more realistic scenario representation.
    \item We propose simple yet efficient baselines leveraged BLIP-2 model. Our proposed methods excel in capturing contextual information surrounding children, achieving strong performance in recognizing both general actions and dangerous situations.
\end{itemize}

\section{Related work}
\subsection{Action Recognition}

% \subsubsection{Convolutional Neural Networks}

Chen et al. \cite{chen2021deep} introduced various models based on \textbf{convolutional neural networks (CNN)} and achieved high accuracy in action recognition. This approach can be divided into two main types: 2D CNN uses 2D filters to process each video frame independently, capturing mainly spatial information but not explicitly modeling temporal relationships. The advantage of 2D CNN is its smaller size and lower computational cost. 3D CNN uses filters to process the video volume, capturing both spatial and temporal information. However, 3D CNNs are larger in size and more computationally expensive than 2D CNNs. Meanwhile, Lin et al. \cite{lin2019tsm} proposed Temporal Shift Module (TSM), capturing spatio-temporal information similar to 3D-CNN models but with computational costs equivalent to 2D-CNNs. Specifically, uni-directional TSM was developed to handle online video processing by only using information from past frames.

CNNs perform well in action recognition. Studies indicate improved accuracy with 3D CNNs compared to 2D CNNs. However, despite their ability to capture both spatial and temporal information, 3D CNNs do not significantly outperform 2D CNNs in terms of accuracy. Research suggests that both 2D CNNs and 3D CNNs exhibit similar behavior regarding the learning of spatio-temporal representations and the transfer of knowledge to new tasks.

% \subsubsection{Recurrent Neural Network Backbone}

On the other hand, \textbf{recurrent neural networks (RNN)} and their variant Long Short-Term Memory (LSTM) have become powerful tools in video analysis, particularly action recognition. RNNs are effective at capturing temporal relationships between frames, while LSTMs are designed to overcome the vanishing gradient problem of RNNs. However, RNNs/LSTMs also have some limitations, such as high computational costs and issues with vanishing/exploding gradients, though LSTM mitigates this to some extent. Some improved methods have been proposed, such as combining CNN and LSTM to reduce computational costs and improve performance. Several works \cite{shi2015convolutional, cho2014learning} introduced advancements in terms of computation and performance.

% \subsubsection{Vision Transformer Backbone}

\textbf{Vision Transformer (ViT)} \cite{dosovitskiy2020image} leveraged the attention mechanism to analyze relationships between parts of an image. However, ViT faces challenges in capturing temporal information and comes with high computational costs. Recent works, such as TimeSformer \cite{bertasius2021space} and ViViT \cite{arnab2021vivit}, have been proposed to address these limitations by modeling spatio-temporal dependencies in videos.

% \subsubsection{Graph Convolutional Backbone}

% \paragraph{Skeleton Graph} is a popular method for recognizing actions from skeletal data. Each node in the graph corresponds to a joint of the body, while edges connecting the nodes represent the physical relationships between the joints. This approach allows the model to learn the relationships between body parts to accurately recognize actions.

\textbf{Graph convolutional networks (GCN) }are the primary tool for analyzing skeletal graphs and recognizing human actions. GCN helps capture the spatial relationships between body parts. Many works \cite{yan2018spatial,li2019actional} used GCN to extract features from skeleton graphs and achieved positive results. However, GCN models struggle to capture complex temporal information from actions. Some studies have employed attention mechanisms or separate streams for spatial and temporal information to enhance the ability to capture temporal features.

\begin{figure}[t!]
    \centering
    \includegraphics[width=0.8\textwidth]{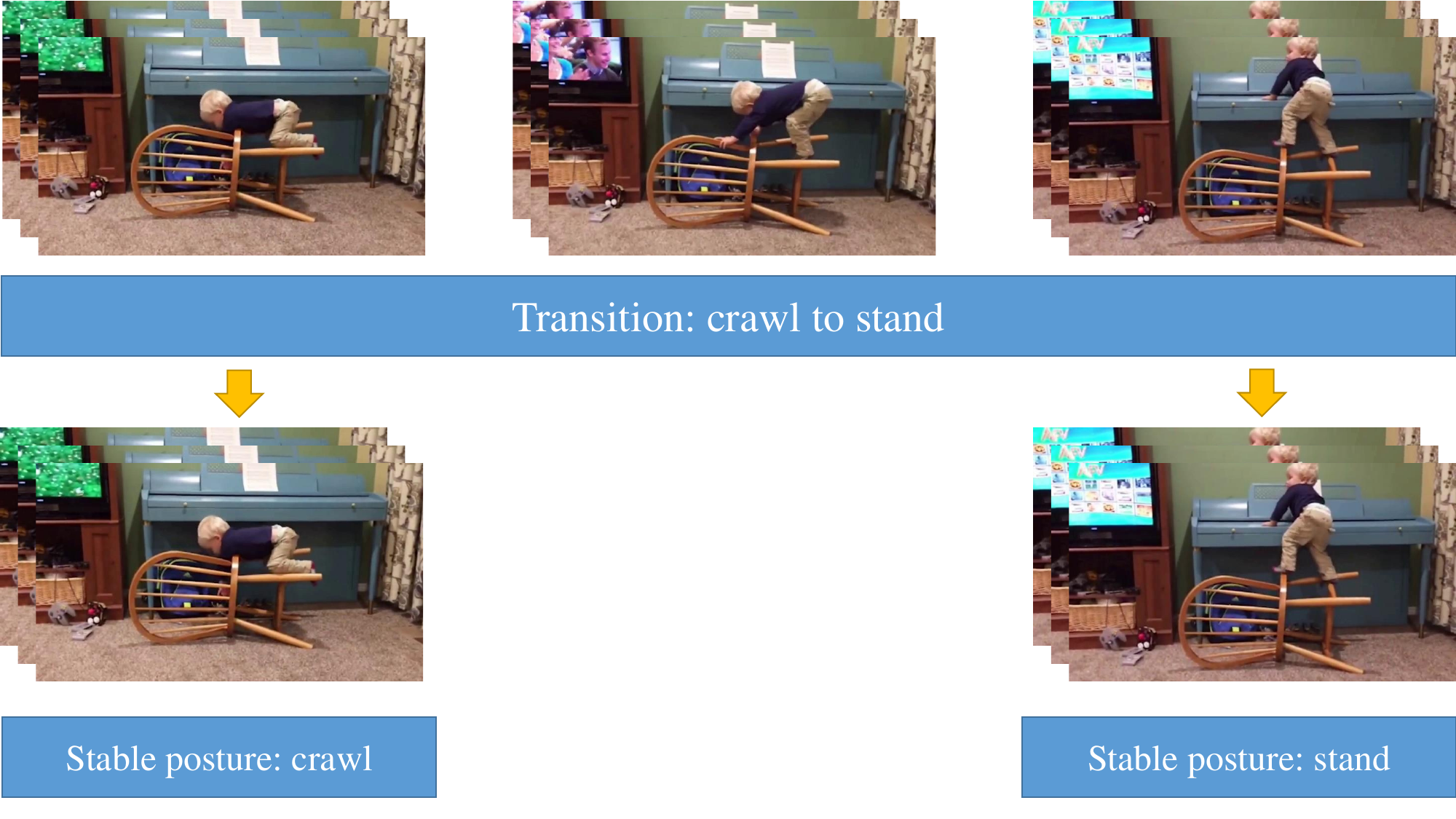}
    \caption{Transition period between crawl posture and stand posture.}
    \captionsetup{justification=centering}
    \label{the_transition_period}
\end{figure}

\subsection{Hazard Recognition}

Research on hazard recognition often focuses on analyzing actions to predict safety risks. Wang et al. \cite{wang2022real} proposed a method for analyzing joint connections between body parts to quickly and accurately identify hazardous actions. However, this method requires a large amount of well-annotated data and does not capture information about the surrounding environment. Meanwhile, Nie et al. \cite{nie2018child} proposed combining action recognition with object detection to assess hazard levels, particularly in children. However, combining both factors requires more complex system processing and additional research to develop more efficient approaches for hazard recognition.

\section{Methodology}
\subsection{Proposed KidRisk Dataset}

To address the challenges in child action recognition and safety detection, we propose a new datasets, including two parts: children's action videos and children's safety images.

\textbf{Children's Action Videos} are developed based on the InfAct dataset \cite{huang2023posture}, which consists of short video clips capturing two actions performed by children with a transition in between, such as sitting, standing, lying down, and crawling. We extend this dataset by extracting and labeling additional video segments from the source, focusing on basic child actions. After processing, the videos are trimmed into shorter clips (up to 5 seconds), each containing only a single action, with the transition periods removed (illustrated by Fig. \ref{the_transition_period}).

\begin{figure}[t!]
    \centering
    \includegraphics[width=\textwidth]{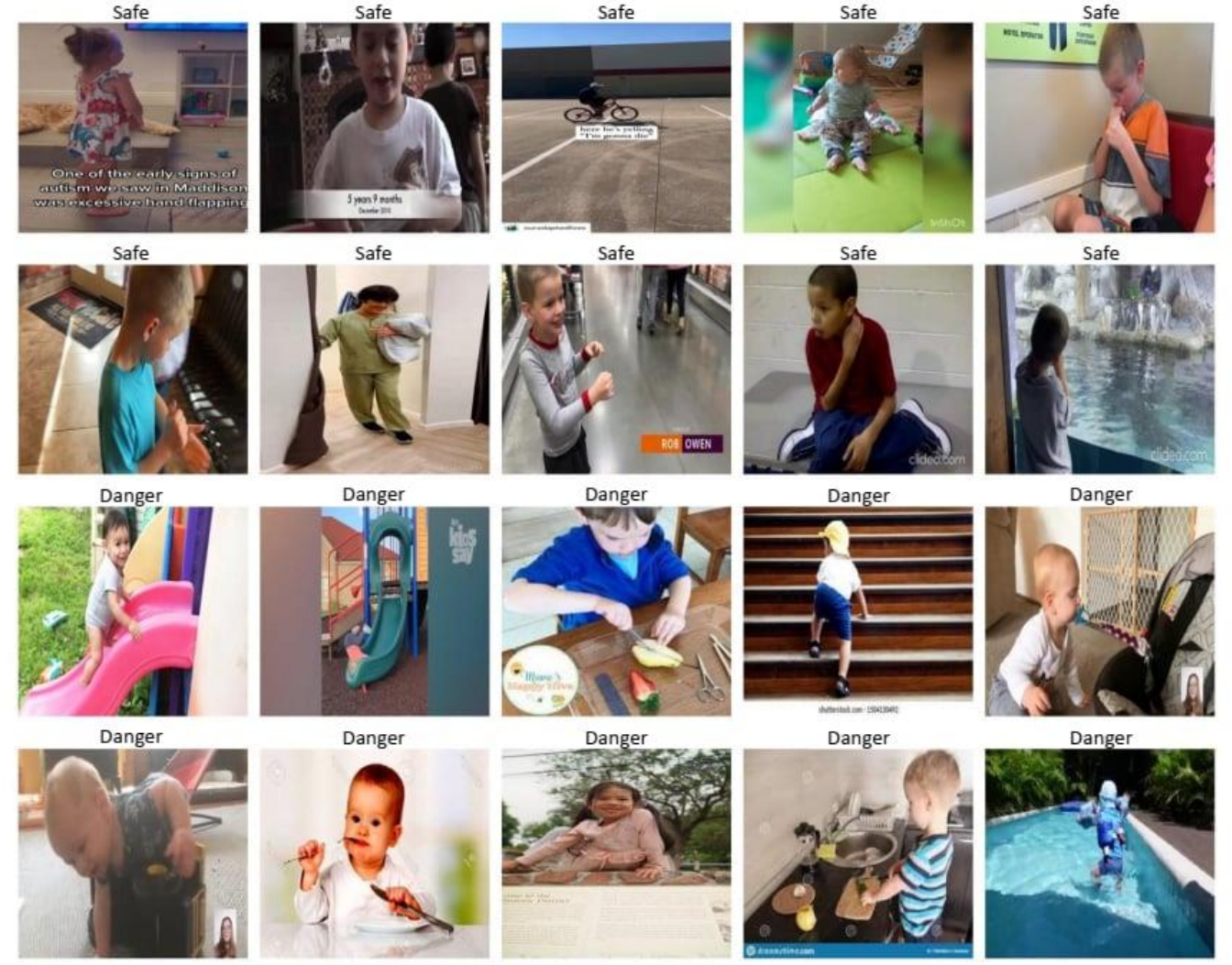}
    \captionsetup{justification=centering}
    \caption{Examples of children's safety images.}
    \label{the_children_dangers_dataset}
\end{figure}

\textbf{Children's Safety Images} are compiled from various sources, including more than 10,000 images depicting children in safe and dangerous situations. These images are categorized into two groups: "Safe" and "Dangerous," with dangerous scenarios including children playing near stairs, handling sharp objects, or being in situations near swimming pools. To increase the dataset's diversity, we supplemented it with additional dangerous situations, creating a rich dataset that accurately reflects the real-world risks children may encounter (see Fig. \ref{the_children_dangers_dataset}).

\subsection{Proposed Baselines}
\begin{figure}[t!]
    \centering
    \begin{subfigure}{0.65\textwidth}
        \centering
        \includegraphics[width=\textwidth]{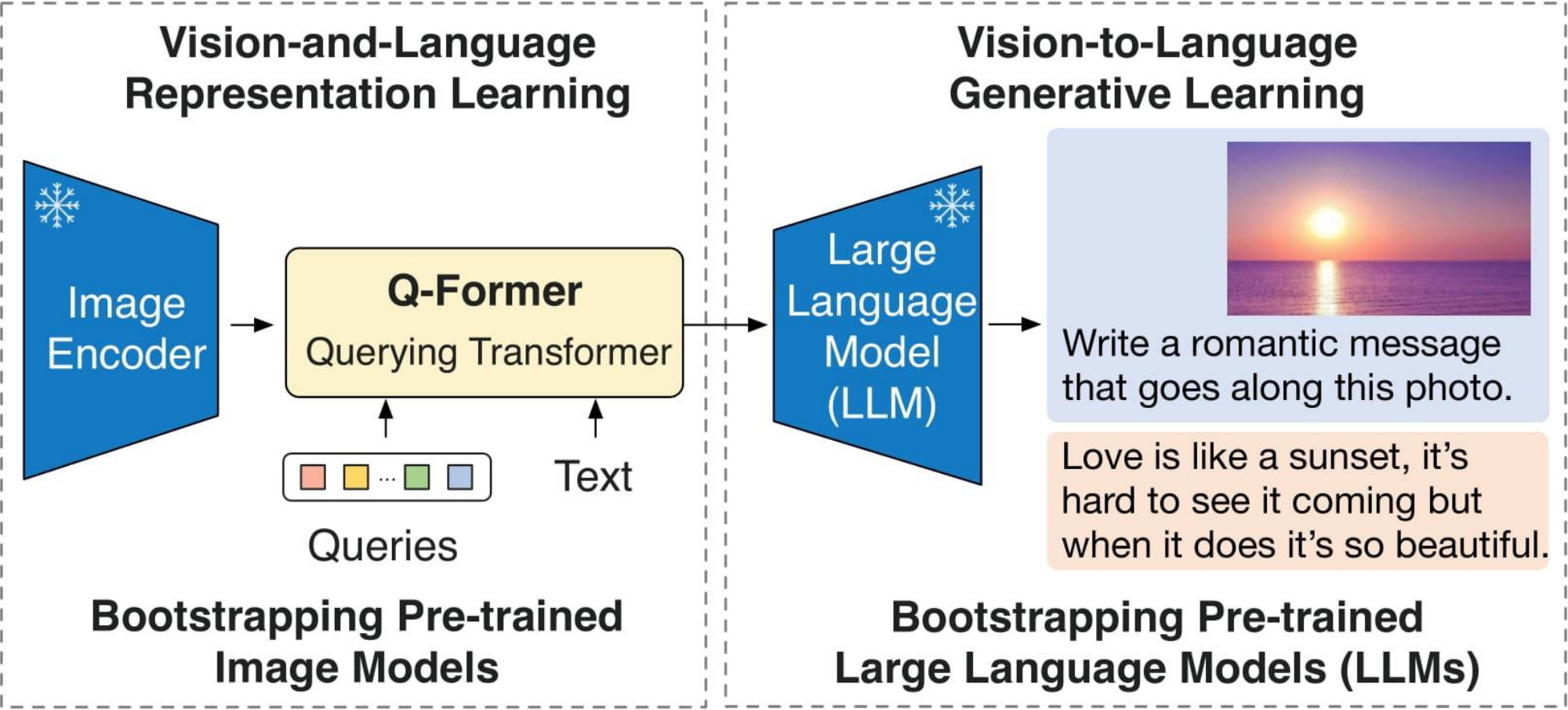}
        \caption{Overview of BLIP-2’s framework.} % Caption cho hình
        \label{fig:BLIP2-framework} % Label cho hình
    \end{subfigure}\hfill
    \begin{subfigure}{0.3\textwidth}
        \centering
        \begin{tabular}{lr}
        \hline
        Models & \makecell{\#Trainable \\ Params}\\ \hline
        BLIP \cite{li2022blip} & 583M \\
        SimVLM \cite{wang2021simvlm} & 1.4B \\
        BEIT-3 \cite{wang2022real} & 1.9B \\
        Flamingo \cite{alayrac2022flamingo} & 10.2B \\
        \hline
        \textbf{BLIP-2 \cite{li2023blip}} & \textbf{188M} \\ \hline
        \end{tabular}
        \caption{Comparison of the number of parameters with previous state-of-the-art models.} % Caption cho bảng
        \label{table:learning-parameter-compare} % Label cho bảng
    \end{subfigure}
    \caption{BLIP-2 framework.}
\end{figure}

The vision-language models have made significant advancements in recent years. However, these models require high computational costs during the training process. In this paper, we utilize the BLIP-2 model, proposed by Li et al. \cite{li2023blip}, to leverage pre-trained components such as a Vision Encoder and a Large Language Model (LLM) with frozen parameters, aiming to reduce computational costs. The Query Transformer (Q-Former) connects information between images and text via a cross-attention mechanism, improving the model’s ability to produce contextually accurate outputs (see Fig. \ref{fig:BLIP2-framework}). The model achieves high performance with only 188M parameters, significantly fewer than SimVLM (1.4B) and Flamingo (10.2B) (see Fig. \ref{table:learning-parameter-compare}).

\subsubsection{Zero-shot Learning Classification} 
\begin{figure}[t!]
    \centering
    \includegraphics[width=0.75\textwidth]{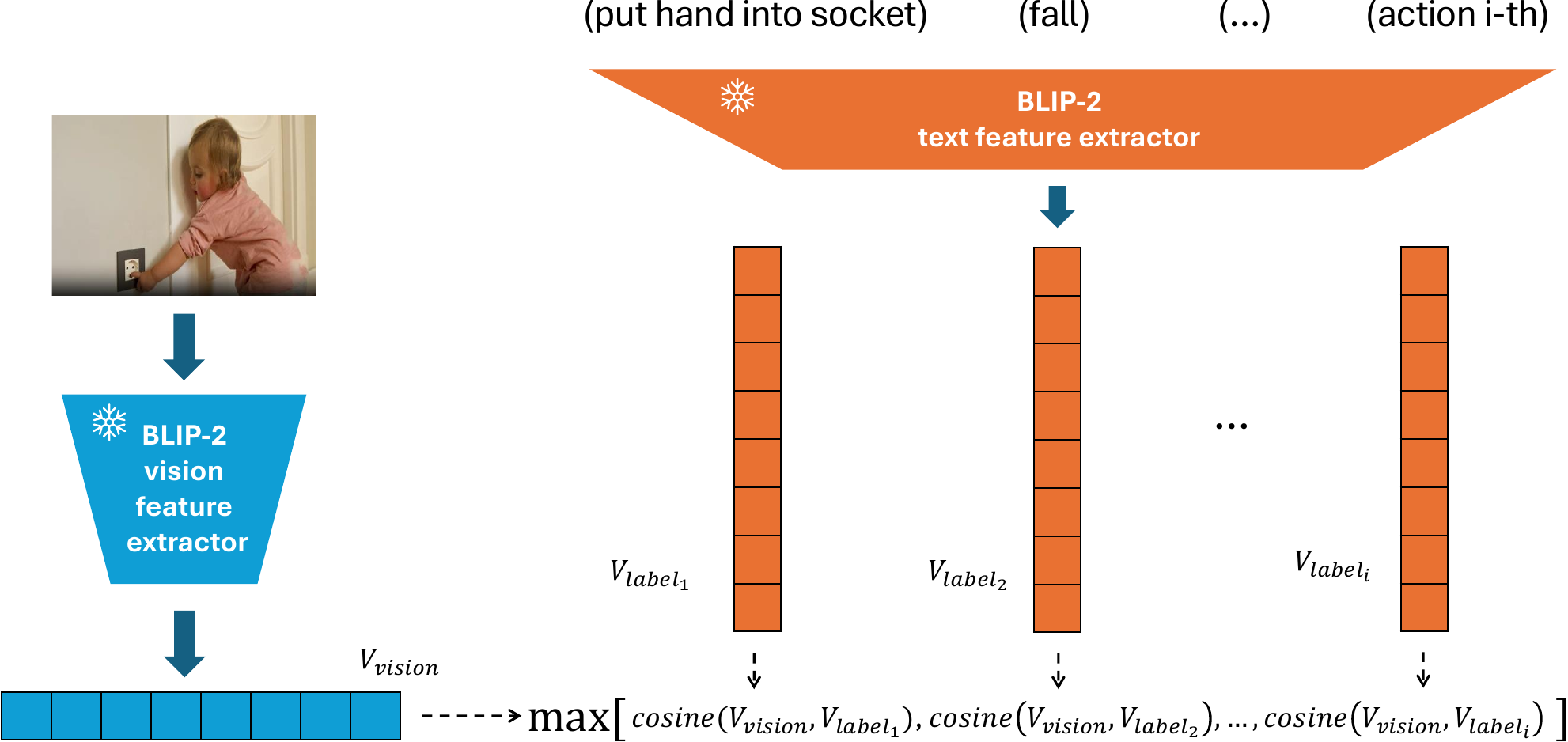}
    \captionsetup{justification=centering}
    \caption{Zero-shot learning for action classification.}
    \label{fig:zeroshot}
\end{figure}

The zero-shot learning method reduces dependence on labeled datasets, making it especially effective when a training set for specific actions is unavailable. In this study, we apply the BLIP-2 model to classify actions in children without prior training (see Fig. \ref{fig:zeroshot}). 

Zero-shot learning leverages the power of pre-trained models to identify actions by comparing the similarity between input images or videos and target action labels, without requiring direct training on the target datasets. The BLIP-2 model uses a Vision Transformer (ViT) to extract features from images/videos. These features are passed through the Q-Former block, which employs a cross-attention mechanism to link visual and textual information, creating feature vectors that represent the visual information. Similarly, action labels or dangerous situations are converted into corresponding feature vectors of the same length using the Q-Former. Cosine similarity (formular \ref{eq:cosine}) is then applied to compare the visual and textual vectors, helping to determine the closest matching action label or dangerous situation based on the highest similarity score:
\begin{equation} \label{eq:cosine} 
    simalarity = \frac{\Vec{V}_{vision}\cdot\Vec{V}_{label}}{||\Vec{V}_{vision}||\cdot||\Vec{V}_{label}||}. 
\end{equation}

\subsubsection{Vision-Language Transfer Learning}
\begin{figure}[t!]
    \centering
    \includegraphics[width=\textwidth]{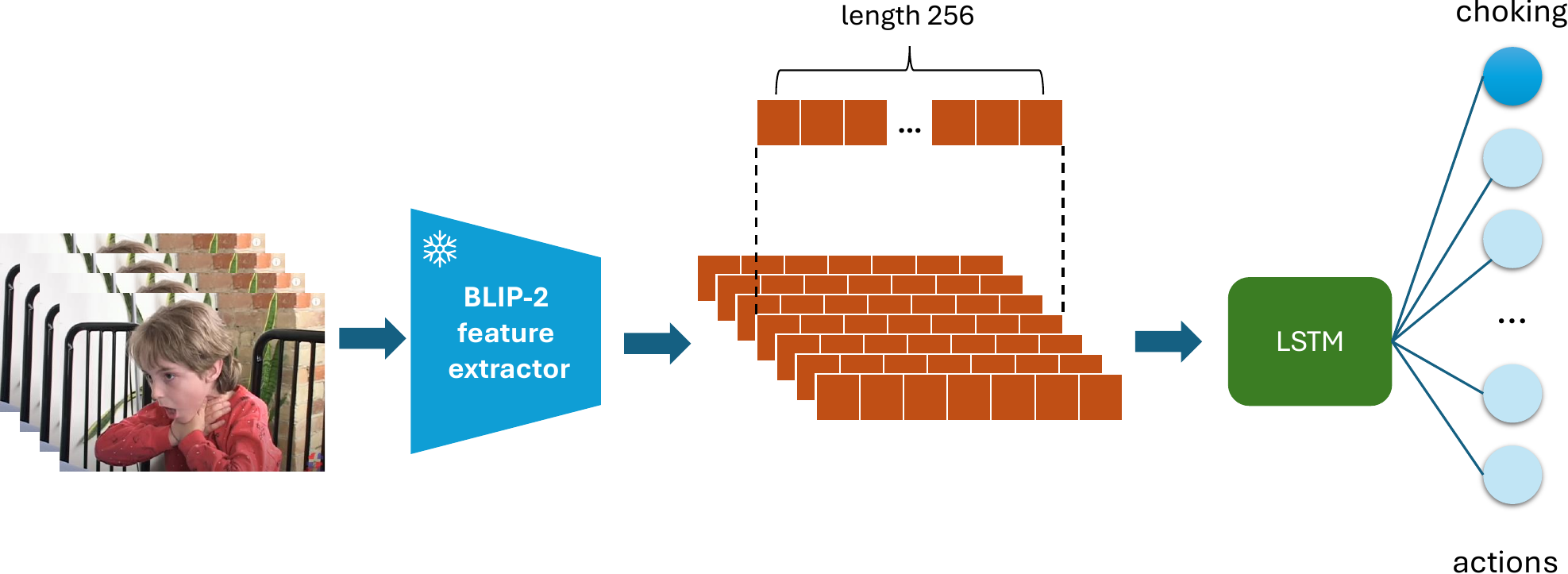}
    \captionsetup{justification=centering}
    \caption{Transfer learning for action recognition.}
    \label{fig:transfer_video}
\end{figure}

In the task of recognizing children's actions in videos, we propose a transfer learning method using BLIP-2 as a feature extractor to obtain feature vectors for each frame (see Fig. \ref{fig:transfer_video}). These vectors are then fed into an LSTM network, which helps the model understand the temporal relationships between frames. Finally, the information is processed through linear layers for action classification. This process not only enhances the accuracy of recognizing children's actions but also reduces the requirement for extensive training data, as the model has been pre-trained on various tasks. This offers significant benefits in improving action recognition capabilities without the need to collect a large dataset.

For danger detection, the study suggests an approach based on processing each frame of the video independently to determine whether it contains a dangerous situation. By applying transfer learning with BLIP-2, the model can extract important features from each frame, which are then input into classification layers with a sigmoid activation function to predict the probability of danger. This approach not only improves the accuracy of detecting dangerous situations but also enhances the safety of children in everyday activities. Monitoring and analyzing each moment allows parents or caregivers to intervene promptly, reducing the risk of accidents.

Our training process involves several key steps to ensure that the model's parameters are optimized for achieving the best performance in action recognition and danger detection tasks. First, the input data undergoes preprocessing. Images from the video are normalized to fit the input format of the BLIP-2 model, typically including resizing to the standard size of (224, 224) and normalizing pixel values. For video data, to reduce load and retain important information, only representative frames are selected from each second of the video for processing. During the training process, the loss function plays a crucial role in guiding the model to optimize its parameters. For the danger detection task, the Binary Cross-Entropy (BCE) loss function is used. This function is suitable for binary classification problems, where it compares the model's predicted probabilities with the actual labels. The BCE loss formula helps the model adjust its parameters so that predictions are as close as possible to the true labels. For the action recognition task, the Cross-Entropy loss function is applied, allowing the model to accurately classify actions across multiple classes. A significant factor in the training process is the issue of overfitting. To address this, L2 regularization is employed. L2 regularization helps mitigate the risk of the model's parameters becoming too large, thereby enhancing the model's ability to generalize to unseen data. The regularization coefficient is adjusted to control the impact of regularization on the loss function.

One notable challenge in training is data imbalance, especially in the case of danger detection. Typically, the number of samples labeled as dangerous is much fewer than those labeled as safe, leading the model to become biased toward the more prevalent class. To overcome this, samples labeled as dangerous are augmented by duplicating them, thus balancing the quantity with safe samples.

The training process runs with a learning rate of \(\alpha = 0.0001\) and the computational resources used include a single T4 GPU, allowing the model to optimize effectively for both tasks.

\section{Experimental Results}

\begin{table}[t!]
    \centering
    \caption{Experimental results of action classification models based on zero-shot learning on the children's action videos.}
    \begin{tabular}{lcc}
    \hline
    Methods  & Backbone & Accuracy\\ 
    \hline
    S3D \cite{miech2020end} & 3D-CNN  & 23.40\% \\
    Alpro \cite{li2022align}  & ViT & 32.10\% \\
    BLIP-2 &  ViT & 62.21\%    \\ 
    \textbf{BLIP-2 + LSTM} & \textbf{ViT} & \textbf{83.5\%} \\ 
    \hline
    \end{tabular}
    \label{tab:zeroshot_action}
\end{table}

\subsection{Children's Action Classification}
The challenge of not having access to a large-scale dataset for children's action recognition highlights the motivation for developing a zero-shot learning approach for classifying children's actions. In this study, we tested several advanced models, including S3D, Alpro, and BLIP-2, to evaluate their effectiveness in classifying children's actions using zero-shot learning. Models utilizing the ViT backbone demonstrated significantly higher performance compared to those based on CNN backbones. Among these, BLIP-2 showed superior capabilities compared to other methods (see Table \ref{tab:zeroshot_action}). 

While action recognition using zero-shot learning with the BLIP-2 model has achieved some impressive results, it cannot yet be considered truly effective in classifying children's actions. Specifically, the performance of this method remains limited, suggesting that the lack of contextual information from training data can hinder the model's ability to accurately recognize complex actions. However, when applying transfer learning, the results obtained are highly promising. Fine-tuning the BLIP-2 model on a specific dataset has led to a significant improvement in classification performance, with accuracy increasing by 21.3\% compared to the previous zero-shot learning method (see Table \ref{tab:zeroshot_action}). This demonstrates that using transfer learning not only allows the model to learn from the features of the target data but also enhances its generalization ability and accuracy in recognizing children's actions.

\begin{table}[t!]
\centering
\caption{Experimental results of transfer learning and other experiments on the danger situation images.}
\label{tab:transfer_danger}
\begin{tabular}{lc}
\hline
Methods                                  & Accuracy \\ \hline
Resnet                                     & 85.1\%       \\
ViT                                          & 75.4\%      \\
BLIP-2 (zero-shot)                            & 56.1\%     \\
\textbf{BLIP-2 + Transfer learning}           & \textbf{96.1\%}\\
\hline
\end{tabular}
\end{table}

\subsection{Children's Danger Recognition}
The pretrained BLIP-2 model has not yet demonstrated effective capabilities in detecting dangerous situations. Specifically, the zero-shot learning method shows relatively low accuracy, while traditional models like ViT and Resnet achieve better results. However, after applying transfer learning, the BLIP-2 model has proven to possess superior detection capabilities, with significantly higher accuracy compared to traditional models on the dataset labeled with dangerous and safe situations (see Table \ref{tab:transfer_danger}).

\begin{figure}[t!]
    \centering
    \includegraphics[width=\textwidth]{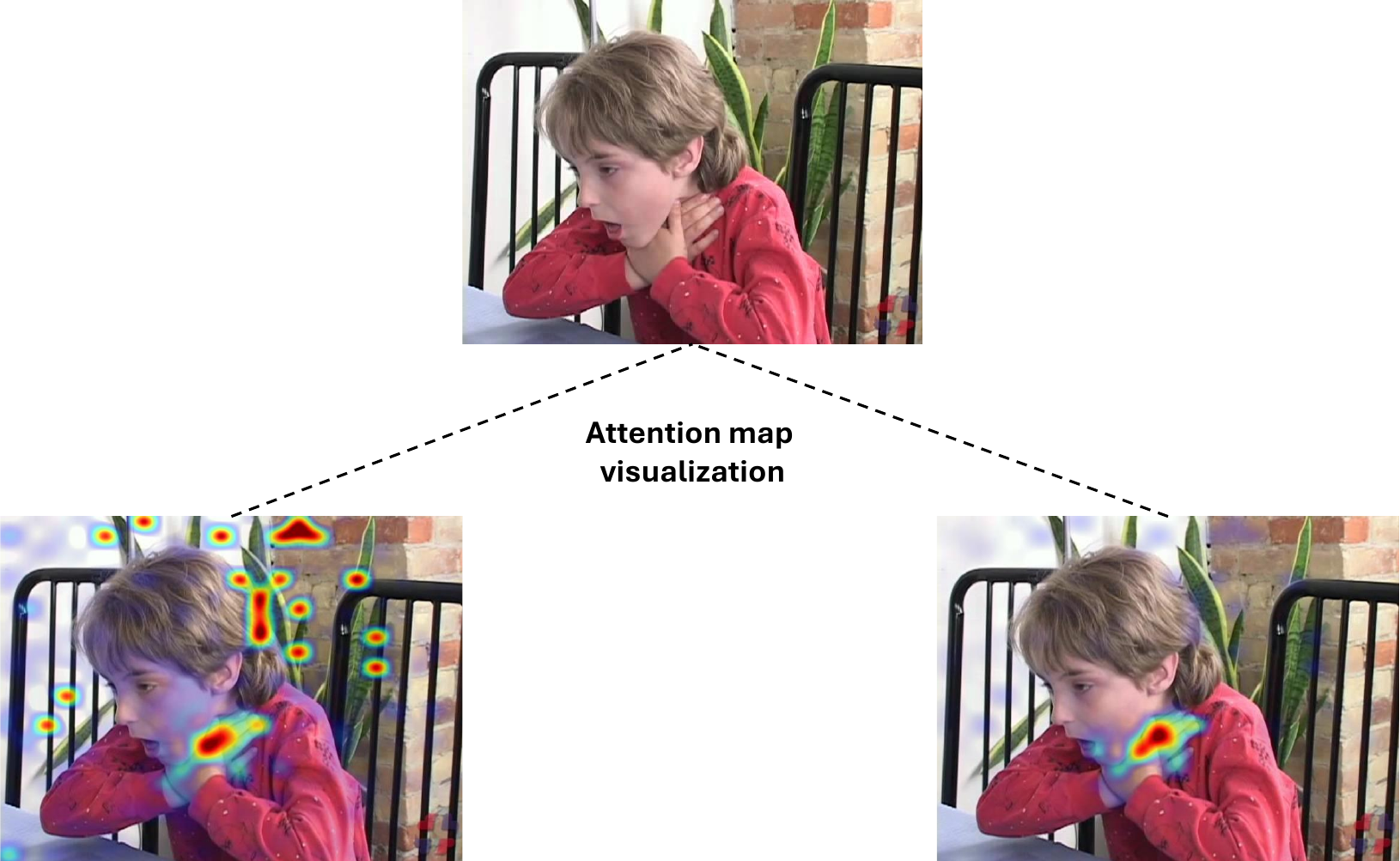}
    \captionsetup{justification=centering}
    \caption{Visualized attention maps of BLIP-2 for action recognition.}
    \label{fig:compare_attention_map}
\end{figure}

We also utilized attention maps to interpret the predictions of the BLIP-2 model (see Fig. \ref{fig:compare_attention_map}), revealing how the model focuses on important regions in the images to detect dangerous situations. The differences between the methods become evident when applying traditional deep learning models, highlighting that leveraging transfer learning has significantly improved the detection capabilities for children's safety in potentially dangerous scenarios.

\section{Conclusion}
In this paper, we introduced the comprehensive KidRisk dataset, encompassing video clips of children's actions and images of hazardous situations, designed to push the boundaries of risk recognition in children's activities. We also develop a simple yet efficient approach for identifying dangerous actions in children through the use of the vision-language BLIP-2 model. Our experimental findings reveal that the integration of BLIP-2 with transfer learning techniques not only delivers exceptional performance but also underscores the potential of vision-language models in real-world applications. Experimental results demonstrated that the combination of BLIP-2 with transfer learning techniques achieved high performance. These results highlight the feasibility of vision-language models in advancing child safety, paving the way for more intelligent, context-aware monitoring systems capable of preemptively identifying and mitigating risks in unsupervised environments.

\section*{Acknowledgement}

This research is supported by research funding from Faculty of Information Technology, University of Science, Vietnam National University - Ho Chi Minh City.

\bibliographystyle{splncs04}
\bibliography{references/refs}
\end{document}